\documentclass{article}
\usepackage{amsmath}
\usepackage{amsfonts}
\usepackage{amssymb}
\usepackage{graphicx}
\usepackage{url}

\begin{document}
\newtheorem{theorem}{Theorem}{\bfseries}{\itshape}
\newtheorem{lemma}{Lemma}{\bfseries}{\itshape}
\newtheorem{corollary}{Corollary}{\bfseries}{\itshape}
\newtheorem{algorisme}{Algorithm}{\bfseries}{\itshape}
\newtheorem{procedure}{Procedure}{\bfseries}{\itshape}
\newtheorem{definition}{Definition}{\bfseries}{\itshape}

\title{Marginality: a numerical mapping
for enhanced treatment of nominal and hierarchical attributes
\thanks{This work was partly supported by
the Government of Catalonia under grant 2009 SGR 1135, by
the Spanish Government 
through projects TSI2007-65406-C03-01
``E-AEGIS'' and CONSOLIDER INGENIO 2010 CSD2007-00004 ``ARES'',
and by the European Comission under FP7 project ``DwB''. 
The author
is partially supported as an ICREA Acad\`emia researcher
by the Government of Catalonia.}
}

\author{Josep Domingo-Ferrer}

\date{Universitat Rovira i Virgili\\ 
Dept. of Computer Engineering and Mathematics\\ 
UNESCO Chair in Data Privacy\\
Av. Pa\"{\i}sos Catalans 26\\
E-43007 Tarragona, Catalonia\\
Tel.: +34 977558270\\
Fax: +34 977559710\\
E-mail josep.domingo@urv.cat}

\maketitle

\begin{abstract}
The purpose of statistical disclosure control (SDC) of microdata, 
a.k.a. data anonymization
or privacy-preserving data mining, is to publish
data sets containing the answers of individual respondents in such
a way that the respondents corresponding to the released records
cannot be re-identified and the released data are analytically useful.
SDC methods are either based on masking the original data,
generating synthetic versions of them or creating hybrid versions
by combining 
original and synthetic data. The choice of SDC methods for categorical
data, especially nominal data, 
is much smaller than the choice of methods for numerical data.
We mitigate this problem by introducing a numerical mapping 
for hierarchical nominal data which allows computing means,
variances and covariances on them. 
\end{abstract}

\section{Introduction}

Statistical disclosure 
control (SDC, \cite{Domi08surv,handbook,Will01,Duncan,Lenz}),
a.k.a. data anonymization and sometimes as privacy-preserving
data mining,
aims at making possible the publication of statistical data
in such a way that the individual responses of specific users
cannot be inferred from the published data and background knowledge
available to intruders. If the data set being published consists of 
records corresponding to individuals, usual SDC methods operate
by masking original data (via perturbation or detail reduction),
by generating synthetic (simulated) data preserving some statistical
features of the original data or by producing hybrid data obtained
as a combination of original and synthetic data.
Whatever the protection method chosen, the resulting data should
still preserve enough analytical validity for their publication
to be useful to potential users.

A microdata set can be defined as a file 
with a number of records, where each record contains a number of 
attributes on an individual respondent. 
Attributes can be classified depending on their range
and the operations that can be performed on them:
\begin{enumerate}
\item {\em Numerical}. An attribute is considered
numerical if arithmetical operations
can be performed on it. Examples are income and age.
When designing methods to protect numerical data,
one has the advantage that
arithmetical operations are possible, and the drawback
that every combination of numerical values in the
original data set is likely to be unique, which leads
to disclosure if no action is taken.
\item {\em Categorical}. An attribute is considered
categorical when it takes values over a finite set
and standard arithmetical operations on it do not make sense.
Two main types of categorical attributes can be distinguished:
\begin{enumerate}
\item {\em Ordinal}. An ordinal attribute takes values
in an ordered range of categories.
Thus, the $\leq$, $\max$
and $\min$ operators are meaningful and can be used
by SDC techniques for ordinal data.
The instruction level and the political
preferences (left-right) are examples of ordinal attributes.
\item {\em Nominal}. A nominal attribute takes
values in an unordered range of categories.
The only possible operator is comparison for
equality. Nominal attributes can further be divided
into two types:
\begin{enumerate}
\item {\em Hierarchical}. A hierarchical nominal attribute
takes values from a hierarchical classification. For example, 
plants are classified using Linnaeus's taxonomy,
the type of a disease is also selected from a hierarchical
taxonomy,
and the type of an attribute
can be selected from the hierarchical classification we propose
in this section.
\item {\em Non-hierarchical}. A non-hierarchical nominal attribute
takes values from a flat hierarchy. Examples of such attributes
could be the preferred soccer team, the address of an individual, 
the civil status (married, single, divorced, widow/er), the eye
color, etc.
\end{enumerate}
\end{enumerate}
\end{enumerate}

This paper focuses on finding a numerical
mapping of nominal attributes, and more precisely
hierarchical nominal attributes. In addition to other 
conceivable applications not dealt with in this paper,
such a mapping can be used to 
anonymize nominal data in ways so far reserved to numerical
data. The interest of this is that many more SDC methods
exist for anonymizing numerical data than categorical 
and especially nominal data. 

Assuming a hierarchy is less restrictive
than it would appear, because very often a 
non-hierarchical attribute can be turned into
a hierarchical one if its flat hierarchy can be developed into 
a multilevel hierarchy. For instance, the preferred soccer
and the address of an individual have been mentioned as 
non-hierarchical attributes; however, a hierarchy of soccer teams
by continent and country could be conceived, and addresses can be
hierarchically clustered by neighborhood, city, state, country, etc.
Furthermore, well-known approaches to anonynimization, like $k$-anonymity
~\cite{Sama01}, assume that any attribute can be generalized,
{\em i.e.} that an attribute hierarchy can be defined and values
at lower levels of the hierarchy can be replaced by values at higher
levels.

\subsection{Contribution and plan of this paper}

We propose to associate a number 
to each categorical value of a hierarchical nominal attribute,
namely a form of centrality
of that category within the attribute's hierarchy. 
We show how this allows 
computation of centroids, variances and covariances of 
hierarchical nominal data.

Section~\ref{back} gives background 
on the variance 
of hierarchical nominal attributes.
Section~\ref{map} defines a tree centrality measure
called marginality and presents
the numerical mapping. Section~\ref{analysis} exploits
the numerical mapping to
compute means, variances and covariances of hierarchical 
nominal data.
Conclusions are drawn in Section~\ref{sec5}.

\section{Background}
\label{back}

We next recall the variance measure for hierarchical
nominal attributes introduced in~\cite{infsci1}. To the best
of our knowledge, this is the first measure which captures
the variability of a sample of values of a hierarchical 
nominal attribute by taking into account the semantics
of the hierarchy. 
The intuitive 
idea is that a set of nominal values belonging to categories which
are all children of the same parent category in the hierarchy
has smaller variance that a set with children from different
parent categories. 

\begin{algorisme}[Nominal variance in~\cite{infsci1}]
\label{enc}~\\
\begin{enumerate}
\item\label{pas1} Let the hierarchy of categories of a nominal
attribute $X$ be such that $b$ is the
maximum number of children that a parent category can have in the
hierarchy.
\item\label{pas2} Given a sample $T_X$ of nominal categories drawn from $X$,
place them in the tree representing the hierarchy of $X$.
Prune the subtrees whose nodes have no associated sample values.
If there are repeated sample values,
there will be several nominal values associated to one or more
nodes (categories) in the pruned tree.
\item\label{pastres} Label as follows the
edges remaining in the tree from the root node to each of its children:
\begin{itemize}
\item If $b$ is odd, consider the following succession of labels
$l_0= (b-1)/2$, $l_1=(b-1)/2-1$, $l_2=(b-1)/2+1$,
$l_3=(b-1)/2-2$, $l_4=(b-1)/2+2$, $\cdots$, $l_{b-2}=0$, $l_{b-1}=b-1$.
\item If $b$ is even, consider the following succession of labels
$l_0=(b-2)/2$, $l_1=(b-2)/2+1$, $l_2=(b-2)/2-1$, $l_3=(b-2)/2+2$,
$l_4=(b-2)/2-2$, $\cdots$, $l_{b-2}=0$, $l_{b-1}=b-1$.
\item Label the
edge leading to the child with most categories associated
to its descendant subtree 
as $l_0$, the edge leading to the child with the second highest number of
categories associated to its 
descendant
subtree as $l_1$, the one leading to the child with the third highest
number of categories associated to its descendant subtree 
as $l_2$ and, in general, the edge leading to the child with the $i$-th
highest number of categories associated to its descendant
subtree as $l_{i-1}$. Since there
are at most $b$ children, the set of labels
$\{l_0, \cdots, l_{b-1} \}$ should
suffice. Thus an {\em edge label} can be viewed as a $b$-ary
digit (to the base $b$).
\end{itemize}
\item\label{pas4} Recursively repeat Step~\ref{pastres} taking instead
of the root node each of the root's child nodes.
\item\label{pas5}
Assign to values
associated to each node in the hierarchy a {\em node label} consisting
of a $b$-ary number
constructed from the edge labels, more specifically as
the concatenation of the $b$-ary digits labeling the
edges along the path from the root to the node:
the label of the edge starting from the root is the most significant one
and the edge label closest to the specific node is the least significant one.
\item\label{pas6} Let $L$ be the maximal length of the leaf $b$-ary labels.
Append as many $l_0$ digits
as needed in the least significant positions to the shorter
labels so that all of them eventually consist of $L$ digits.
\item\label{pas7} Let $T_X(0)$ be the set of $b$-ary digits in the least
significant positions of the node labels (the ``units'' positions);
let $T_X(1)$ be the set
of $b$-ary digits in the second least significant positions
of the node labels (the ``tens'' positions), and so on, until
$T_X(L-1)$ which is the set of digits in the most significant
positions of the node labels.
\item\label{pas8} Compute the variance of the sample as
\[ Var_H(T_X) = Var(T_X(0))+ b^2 \cdot Var(T_X(1)) + \cdots \]
\begin{equation}
\label{nomvar}
+ b^{2(L-1)} \cdot Var(T_X(L-1)) 
\end{equation}
where $Var(\cdot)$ is the usual numerical variance.
\end{enumerate}
\end{algorisme}

In Section~\ref{vari} below we will show
that an equivalent measure can be obtained in a 
simpler and more manageable way.

\section{A numerical mapping for nominal hierarchical data}
\label{map}

Consider a nominal attribute $X$ taking values from a hierarchical
classification. Let $T_X$ be a sample of values of $X$.
Each value $x \in T_X$ can be associated two numerical values:
\begin{itemize}
\item The sample frequency of $x$;
\item Some centrality measure  of $x$ within the hierarchy
of $X$.
\end{itemize}

While the frequency depends on the particular sample, centrality 
measures depend both on the attribute hierarchy and the sample. 
Known tree centralities attempt to determine the ``middle'' of 
a tree~\cite{Reid}. We are rather interested in finding how far
from the middle is each node of the tree, that is, how marginal
it is. We next propose an
algorithm to compute a new measure of the marginality
of the values in the sample $T_X$. 

\begin{algorisme}[Marginality of nominal values]
\label{marg}~\\
\begin{enumerate}
\item Given a sample $T_X$ of nominal categorical values drawn from 
$X$, place them in the tree representing the hierarchy of $X$.
There is a one-to-one mapping between the set of tree nodes
and the set of categories where $X$ takes values.
Prune the subtrees whose nodes have no associated sample values.
If there are repeated sample values,
there will be several nominal values associated to one or more
nodes (categories) in the pruned tree.
\item Let $L$ be the depth of the pruned tree. 
Associate weight $2^{L-1}$ to edges linking the root of the 
hierarchy to its immediate descendants (depth 1), 
weight $2^{L-2}$ to edges linking the depth 1 descendants to 
their own descendants (depth 2), and so on, up to weight $2^0=1$ 
to the edges linking descendants at depth $L-1$ with those at depth $L$.
In general, weight $2^{L-i}$ is assigned to edges linking nodes
at depth $i-1$ with those at depth $i$, for $i=1$ to $L$.
\item For each nominal value $x_j$ in the sample,
its marginality $m(x_j)$ is defined and computed as 
\[ m(x_j) = \sum_{x_l \in T_X -\{x_j\}} d(x_j,x_l) \]
where $d(x_j,x_l)$
is the sum of the edge weights along the shortest 
path from the tree node corresponding
to $x_j$ and the tree node corresponding to $x_l$.
\end{enumerate}
\end{algorisme}

Clearly, the greater $m(x_j)$, the more marginal 
({\em i.e.} the less central) is $x_j$. 
Some properties follow which illustrate the rationale
of the distance and the weights 
used to compute the marginality. 

\begin{lemma}
\label{distd}
$d(\cdot,\cdot)$ is a distance in the mathematical sense.
\end{lemma}

Being the length of a path,
it is immediate to check that $d(\cdot,\cdot)$ satisfies 
reflexivity, symmetry and subadditivity.
The rationale of the above exponential weight scheme is to give 
more weight to differences at higher levels of the hierarchy; 
specifically, the following property is satisfied.

\begin{lemma}
\label{lemgreat}
The distance between any non-root node $n_j$ and its 
immediate ancestor is greater
than the distance between $n_j$ and any of its descendants. 
\end{lemma}

{\bf Proof:} 
Let $L$ be the depth of the overall tree and $L_j$ be the depth of $n_j$.
The distance between 
$n_j$ and its immediate ancestor is $2^{L-L_j}$. 
The distance between $n_j$ and its most distant ancestor is
\[ 1+2+ \cdots+ 2^{L-L_j-1}=2^{L-L_j}-1 \]
$\Box$

\begin{lemma}
The distance between any two nodes at the same
depth is greater than the longest distance within
the subtree rooted at each node.
\end{lemma}

{\bf Proof:} 
Let $L$ be the depth of the overall tree and
$L_j$ be the depth of the two nodes. 
The shortest distance between both nodes occurs when they
have the same parent and it is
\[ 2 \cdot 2^{L-L_j} = 2^{L-L_j+1}.\]
The longest distance within any of the two subtrees
rooted at the two nodes at depth $L_j$ is the length
of the path between
two leaves at depth $L$, which is
\[ 2\cdot(1+2+ \cdots+ 2^{L-L_j-1})=2(2^{L-L_j}-1)=2^{L-L_j+1}-2 \]
$\Box$

\section{Statistical analysis of numerically mapped nominal data}
\label{analysis}

In the previous section we have shown how a nominal value $x_j$ can be 
associated a marginality measure $m(x_j)$.
In this section, we show how this numerical magnitude 
can be used in statistical analysis.

\subsection{Mean}

The mean of a sample of nominal values cannot be computed in the 
standard sense. However, it can be reasonably approximated by
the least marginal value, that is, by the most central value
in terms of the hierarchy.

\begin{definition}[Marginality-based approximated mean]
\label{margmean}
Given a sample $T_X$ of a hierarchical nominal attribute $X$, 
the marginality-based approximated mean is defined as
\[ Mean_M(T_X) = \arg\min_{x_j \in T_X} m(x_j) \] if one wants 
the mean to be a nominal value, or
\[ Num\_mean_M(T_X) = \min_{x_j \in T_X} m(x_j)\]
if one wants a numerical mean value.
\end{definition}

\subsection{Variance}
\label{vari}

In Section~\ref{back} above, we recalled a measure
of variance of a hierarchical nominal attribute proposed
in~\cite{infsci1} which 
takes the semantics of the hierarchy into account. 
Interestingly, it turns out that the average marginality of 
a sample is an equivalent way to capture the same notion of variance.

\begin{definition}[Marginality-based variance]
\label{defvar}
Given a sample $T_X$ of $n$ values
drawn from a hierarchical nominal attribute $X$, 
the marginality-based sample variance is defined as
\[ Var_M(T_X) = \frac{\sum_{x_j \in T_X} m(x_j)}{n} \]
\end{definition}

The following lemma is proven in the Appendix.

\begin{lemma}
\label{equiv}
The $Var_M(\cdot)$ measure and 
the $Var_H(\cdot)$ specified by Algorithm~\ref{enc} 
in Section~\ref{back} are equivalent.
\end{lemma}

\subsection{Covariance matrix}

It is not difficult to generalize the sample variance
introduced in Definition~\ref{defvar} 
to define the sample covariance of two nominal attributes.

\begin{definition}[Marginality-based covariance]
\label{defcovar}
Given a bivariate sample $T_{(X,Y)}$ consisting of $n$ ordered pairs of
values $\{(x_1,y_1), \cdots, (x_n,y_n)\}$ 
drawn from the ordered pair of nominal attributes $(X,Y)$, 
the marginality-based sample covariance is defined as
\[ Covar_M(T_{(X,Y)}) = \frac{\sum_{j=1}^n \sqrt{m(x_j)m(y_j)}}{n} \]
\end{definition}

The above definition yields a non-negative covariance whose 
value is higher when the marginalities of the values taken by $X$ and $Y$
are positively correlated: as the values taken by $X$ become more
marginal, so become the values taken by $Y$.

Given a multivariate data set $T$ containing a sample of $d$ nominal attributes
$X^1, \cdots, X^d$,
using Definitions~\ref{defvar} and~\ref{defcovar} yields
a covariance matrix ${\bf S} = \{ s_{jl} \}$, for $1 \leq j \leq d$ 
and $1 \leq l \leq d$, where $s_{jj} = Var_M(T_j)$, 
$s_{jl}=Covar_M(T_{jl})$ for $j \neq l$, $T_j$ is the column of values taken
by $X^j$ in $T$ and $T_{jl}=(T_j,T_l)$.

We can use the following distance definition
for records with numerical, nominal or hierarchical attributes.

\begin{definition}[SSE-distance]
The SSE-distance between two records
${\bf x}_1$ and ${\bf x}_2$ in a data set
with $d$ attributes is
\begin{equation}
\label{ssedist}
\delta({\bf x}_1,{\bf x}_2) = 
\sqrt{\frac{(S^2)^1_{12}}{(S^2)^1} 
+ \cdots + \frac{(S^2)^d_{12}}{(S^2)^d}}
\end{equation}
where $(S^2)^l_{12}$ is the variance of the $l$-th attribute
over the group formed by ${\bf x}_1$ and ${\bf x}_2$, and 
$(S^2)^l$ is the variance of the $l$-th attribute over
the entire data set.
\end{definition}

We prove in the Appendix the 
following two theorems stating that the distance above satisfies
the properties of a mathematical distance.

\begin{theorem}
\label{teo1}
The SSE-distance on multivariate records
consisting of nominal attributes
based on the nominal variance computed as per
Definition~\ref{defvar}
is a distance in the mathematical sense.
\end{theorem}

\begin{theorem}
\label{teo2}
The SSE-distance on multivariate 
records consisting of ordinal or numerical attributes
based on the usual numerical variance 
is a distance in the mathematical sense.
\end{theorem}

By combining the proofs of Theorems~\ref{teo1} and~\ref{teo2}, 
the next corollary follows.

\begin{corollary}
\label{cor1}
The SSE-distance on multivariate records
consisting of attributes of any type, where 
the nominal variance is used for nominal attributes
and the usual numerical variance is used
for ordinal and numerical attributes,
is a distance in the mathematical sense.
\end{corollary}

\section{Conclusions}
\label{sec5}

We have presented a centrality-based
mapping of hierarchical nominal data to numbers.
We have shown how such a 
numerical mapping allows computing means, variances
and covariances of nominal attributes, and distances
between records containing any kind of attributes. 
Such enhanced flexility of manipulation of nominal 
attributes can be used, {\em e.g.} to adapt anonymization
methods intented for numerical data to the treament
of nominal and hierarchical attributes. The only requirement
is that, whatever the treatment, it should not modify the numerical values 
assigned by marginality, in order for the numerical 
mapping to be reversible and allow recovering
the original nominal values after treatment.

\appendix

\section*{Appendix}

{\bf Proof (Lemma~\ref{equiv}):} We will show that, given two samples 
$T_X=\{x_1,\cdots,x_n\}$ and 
$T'_X=\{x'_1,\cdots,x'_n\}$ of a nominal attribute $X$, both 
with the same cardinality $n$, it holds that
$Var_M(T_X) < Var_M(T'_X)$ if and only if $Var_H(T_X) < Var_H(T'_X)$.

Assume that $Var_M(T_X) < Var_M(T'_X)$. Since both samples have the same
cardinality, this is equivalent to
\[ \sum_{j=1}^n m(x_j) < \sum_{j=1}^n m(x'_j) \]
By developing the marginalities, we obtain
\[\sum_{j=1}^n \sum_{x_l \in T_X -\{x_j\}} d(x_j,x_l) 
< \sum_{j=1}^n \sum_{x'_l \in T'_X -\{x'_j\}} d(x'_j,x'_l)\]
Since distances are sums of powers of 2, from 1 to $2^{L-1}$,
we can write the above inequality as
\begin{equation}
\label{ineq}
d_0+ 2d_1 + \cdots+ 2^{L-1}d_{L-1} < d'_0+ 2d'_1 + \cdots + 2^{L-1} d'_{L-1}
\end{equation}
By viewing $d_{L-1} \cdots d_1 d_0$ and $d'_{L-1} \cdots d'_1 d'_0$ as binary
numbers, it is easy to see that Inequality (\ref{ineq}) implies 
that some $i$ must exist such that $d_i < d'_i$ and 
$d_{\hat{i}} \leq d'_{\hat{i}}$ for $i < \hat{i} \leq L-1$.
This implies that there are less high-level edge differences associated
to the values of $T_X$ than to the values of $T'_X$. Hence, in terms
of $Var_H(\cdot)$, we have that $Var(T_X(i)) < Var(T'_X(i))$
and $Var(T_X(\hat{i})) \leq Var(T'_X(\hat{i})$ for $i < \hat{i} \leq L-1$.
This yields $Var_H(T_X) < Var_H(T'_X)$.

If we now assume $Var_H(T_X) < Var_H(T'_X)$ we can prove 
$Var_M(T_X) < Var_M(T'_X)$ by reversing the above argument. $\hfill \Box$.

\begin{lemma}
\label{lem1}
Given non-negative $A, A', A'', B, B', B''$ such that
$\sqrt{A} \leq \sqrt{A'} + \sqrt{A''}$ and
$\sqrt{B} \leq \sqrt{B'} + \sqrt{B''}$ it holds that
\begin{equation}
\label{exp1}
\sqrt{A+B} \leq \sqrt{A'+B'} + \sqrt{A''+B''}
\end{equation}
\end{lemma}

{\bf Proof (Lemma~\ref{lem1}): }
Squaring the two inequalities in the lemma assumption, we obtain
\[ A \leq (\sqrt{A'} + \sqrt{A''})^2 \]
\[ B \leq (\sqrt{B'} + \sqrt{B''})^2 \]
Adding both expressions above, we get the square of the left-hand
side of Expression (\ref{exp1})
\[ A + B \leq (\sqrt{A'} + \sqrt{A''})^2 + (\sqrt{B'}+\sqrt{B''})^2 \]
\begin{equation}
\label{exp2}
= A' + A'' + B'+B'' + 2 (\sqrt{A'A''} + \sqrt{B'B''})
\end{equation}
Squaring the right-hand side of Expression (\ref{exp1}), we get 
\[ (\sqrt{A'+B'} + \sqrt{A''+B''})^2 \]
\begin{equation}
\label{exp3}
= A'+B'+A''+B''+2\sqrt{(A'+B')(A''+B'')}
\end{equation}
Since Expressions (\ref{exp2}) and (\ref{exp3}) both contain
the terms $A'+B'+A''+B''$, we can neglect them. Proving Inequality (\ref{exp1})
is equivalent to proving
\[ \sqrt{A'A''} + \sqrt{B'B''} \leq \sqrt{(A'+B')(A''+B'')} \]
Suppose the opposite, that is,
\begin{equation}
\label{exp5}
\sqrt{A'A''} + \sqrt{B'B''} > \sqrt{(A'+B')(A''+B'')} 
\end{equation}
Square both sides:
\[ A'A''+B'B''+2\sqrt{A'A''B'B''} > \]
\[ (A'+B')(A''+B'') = A'A''+B'B''+A'B''+B'A''\]
Subtract $A'A''+B'B''$ from both sides to obtain
\[ 2\sqrt{A'A''B'B''} > A'B''+B'A''\]
which can be rewritten as
\[ (\sqrt{A'B''}-\sqrt{B'A''})^2 < 0\]
Since a real square cannot be negative, 
the assumption in Expression (\ref{exp5}) is false
and the lemma follows. $\Box$

\rule{0cm}{0.5cm}\\
{\bf Proof (Theorem~\ref{teo1}): } We must prove that the SSE-distance is non-negative,
reflexive, symmetrical and subadditive ({\em i.e.} it satisfies
the triangle inequality). 

{\em Non-negativity}. The SSE-distance is defined as a non-negative
square root, hence it cannot be negative.

{\em Reflexivity}. If ${\bf x}_1 = {\bf x}_2$, then 
$\delta({\bf x}_1,{\bf x}_2)=0$. Conversely, if 
$\delta({\bf x}_2,{\bf x}_2)=0$, the variances are all zero,
hence ${\bf x}_1={\bf x}_2$.

{\em Symmetry}. It follows from the definition of the SSE-distance.

{\em Subadditivity.} Given three records ${\bf x}_1$, ${\bf x}_2$
and ${\bf x}_3$, we must check whether
\[\delta({\bf x}_1,{\bf x}_3) \stackrel{?}{\leq} \delta({\bf x}_1,{\bf x}_2)
+ \delta({\bf x}_2, {\bf x}_3) \]
By expanding the above expression using Expression (\ref{ssedist}), we obtain
\[\sqrt{\frac{(S^2)^1_{13}}{(S^2)^1} 
+ \cdots + \frac{(S^2)^d_{13}}{(S^2)^d}} \stackrel{?}{\leq} \]
\begin{equation}
\label{subadd}
\sqrt{\frac{(S^2)^1_{12}}{(S^2)^1} 
+ \cdots + \frac{(S^2)^d_{12}}{(S^2)^d}} + 
\sqrt{\frac{(S^2)^1_{23}}{(S^2)^1} 
+ \cdots + \frac{(S^2)^d_{23}}{(S^2)^d}}
\end{equation}
Let us start with the case $d=1$, that is, with a single attribute, {\em i.e.}
${\bf x}_i = x_i$ for $i=1,2,3$.
To check Inequality (\ref{subadd}) with $d=1$,
we can ignore the variance in the denominators (it is the same
on both sides) and we just need to check
\begin{equation}
\label{subadd2}
\sqrt{S^2_{13}} \stackrel{?}{\leq} \sqrt{S^2_{12}} + \sqrt{S^2_{23}}
\end{equation}
We have
\[ S^2_{13}=Var(\{x_1,x_3\})=\frac{m(x_1)+m(x_3)}{2} \]
\begin{equation}
\label{expexp}
= \frac{d(x_1,x_3)}{2}+\frac{d(x_3,x_1)}{2} = d(x_1,x_3)
\end{equation}
Similarly $S^2_{12}=d(x_1,x_2)$ and $S^2_{23}=d(x_2,x_3)$.
Therefore, Expression (\ref{subadd2}) is equivalent to 
subaddivitity for $d(\cdot,\cdot)$ and the latter holds by 
Lemma~\ref{distd}.
Let us now make the induction hypothesis for $d-1$ 
and prove subadditivity for any $d$. Call now
\[ A:= \frac{(S^2)^1_{13}}{(S^2)^1} 
+ \cdots + \frac{(S^2)^{d-1}_{13}}{(S^2)^{d-1}} \]
\[ A':= \frac{(S^2)^1_{12}}{(S^2)^1} 
+ \cdots + \frac{(S^2)^{d-1}_{12}}{(S^2)^{d-1}} \]
\[ A'':= \frac{(S^2)^1_{23}}{(S^2)^1} 
+ \cdots + \frac{(S^2)^{d-1}_{23}}{(S^2)^{d-1}} \]
\[B:= \frac{(S^2)^d_{13}}{(S^2)^d}; \,\,
 B':= \frac{(S^2)^d_{12}}{(S^2)^d}; \,\,
B'':= \frac{(S^2)^d_{23}}{(S^2)^d} \]
Subadditivity for $d$ amounts to checking whether
\begin{equation}
\label{subadd3}
\sqrt{A+B} \stackrel{?}{\leq} \sqrt{A'+B'} + \sqrt{A''+B''} 
\end{equation}
which holds by Lemma~\ref{lem1} because,
by the induction hypothesis for $d-1$, we have
$\sqrt{A} \leq \sqrt{A'} + \sqrt{A''}$ and,
by the proof for $d=1$,
we have $\sqrt{B} \leq \sqrt{B'} + \sqrt{B''}$.
$\Box$

\rule{0cm}{0.5cm}\\
{\bf Proof (Theorem~\ref{teo2}):} Non-negativity, reflexivity and symmetry
are proven in a way analogous as in Theorem~\ref{teo1}.
As to subaddivity, we just need to prove the case $d=1$,
that is, 
the inequality analogous to Expression (\ref{subadd2}) 
for numerical variances. The proof for general $d$ is 
the same as in Theorem~\ref{teo1}. For $d=1$, we have
\[ S^2_{13} = \frac{(x_1 - x_3)^2}{2}; \,\,
S^2_{12} = \frac{(x_1 - x_2)^2}{2}; \,\,
S^2_{23} = \frac{(x_2 - x_3)^2}{2} \]
Therefore, Expression (\ref{subadd2}) obviously holds
with equality in the case of numerical variances because
\[ \sqrt{S^2_{13}}= 
\frac{x_1 - x_3}{\sqrt{2}} = \frac{(x_1 - x_2) + (x_2 -x_3)}{\sqrt{2}}=
\sqrt{S^2_{12}} + \sqrt{S^2_{23}}\] $\Box$

\section*{Acknowledgments and disclaimer}

Thanks go to Klara Stokes for useful help. 
The authors are with the UNESCO Chair in Data Privacy, 
but they are solely responsible for the views
expressed in this paper, which do not necessarily
reflect the position of UNESCO nor commit that organization.

\bibliographystyle{plain}

\end{document}